\documentclass[10pt,twocolumn,letterpaper]{article}

\usepackage[pagenumbers]{cvpr} 
\usepackage{bm}
\usepackage{multirow}
\usepackage{graphicx}
\usepackage{bm}
\usepackage{algorithm}
\usepackage{algorithmic}
\usepackage{amsmath}
\usepackage{amssymb}
\usepackage{multirow}
\usepackage{xcolor}  
\definecolor{cvprblue}{rgb}{0.21,0.49,0.74}
\usepackage[pagebackref,breaklinks,colorlinks,allcolors=cvprblue]{hyperref}

\def\paperID{4860} 
\def\confName{CVPR}
\def\confYear{2025}

\title{MCCD: Multi-Agent Collaboration-based Compositional Diffusion for \\ 
Complex Text-to-Image Generation}

\author{Mingcheng Li$^{1,2}$$\textsuperscript{*}$$\quad$
        Xiaolu Hou$^{1,2}$$\textsuperscript{*}$$\quad$
        Ziyang Liu$^{3}\quad$
        Dingkang Yang$^{1,2}\footnotemark[4]\quad$ 
        Ziyun Qian$^{1,2}\quad$ \\
        Jiawei Chen$^{1,2}\quad$ 
        Jinjie Wei$^{1,2}\quad$ 
        Yue Jiang$^{1,2}\quad$ 
        Qingyao Xu$^{1,2}\quad$ 
        Lihua Zhang$^{1,2,4,5}$\footnotemark[4] \\ 
        \small$^1$Academy for Engineering and Technology, Fudan University$\,$ 
        \small$^2$Cognition and Intelligent Technology Laboratory (CIT Lab)\\
        \small$^3$School of Future Science and Engineering, Soochow University, Suzhou, China \\
        \small$^4$Jilin Provincial Key Laboratory of
Intelligence Science and Engineering, Changchun, China\\
\small$^5$Engineering Research Center of AI and Robotics, Ministry of Education, Shanghai, China\\
{\tt\small \{mingchengli21, xlhou23\}@m.fudan.edu.cn, dkyang20@fudan.edu.cn}
}

\begin{document}
\maketitle
\renewcommand{\thefootnote}{\fnsymbol{footnote}}
\footnotetext[4]{Corresponding authors. \textsuperscript{*}Equal contributions.} 

\begin{abstract}
Diffusion models have shown excellent performance in text-to-image generation. Nevertheless, existing methods often suffer from performance bottlenecks when handling complex prompts that involve multiple objects, characteristics, and relations. 
Therefore, we propose a Multi-agent Collaboration-based Compositional Diffusion (MCCD) for text-to-image generation for complex scenes.
Specifically, we design a multi-agent collaboration-based scene parsing module that generates an agent system comprising multiple agents with distinct tasks, utilizing MLLMs to extract various scene elements effectively.
In addition, Hierarchical Compositional diffusion utilizes a Gaussian mask and filtering to refine bounding box regions and enhance objects through region enhancement, resulting in the accurate and high-fidelity generation of complex scenes.
Comprehensive experiments demonstrate that our MCCD significantly improves the performance of the baseline models in a training-free manner, providing a substantial advantage in complex scene generation.

\end{abstract}

\section{Introduction}
Recently, diffusion models \cite{dhariwal2021diffusion, yang2023diffusion, sohl2015deep, rombach2022high} have shown significant advancements in Text-to-Image (T2I) generation, such as Stable Diffusion \cite{rombach2022high}, Imagen \cite{saharia2022photorealistic} and DALL-E 2/3 \cite{ramesh2022hierarchical, betker2023improving}.
 However,  despite their noteworthy performance in generating realistic images consistent with text prompts, these models have large limitations in processing complex textual prompts and generating complex scenes, leading to unsatisfactory image generation \cite{feng2022training, liu2022compositional, lian2023llm}.
 Therefore, T2I models require powerful spatial perceptions that can precisely align multiple objects with different attributes and complex relationships involved in compositional prompts.

Some studies attempt to introduce additional conditions to solve the problems above, which can be divided into two parts: (i) spatial information-based methods \cite{li2023gligen, yang2023reco, qu2023layoutllm, zhang2024realcompo, lian2023llm}, and (ii) feedback-based methods. 
Spatial information-based methods utilize additional spatial information (e.g., layouts and boxes) as conditions to enhance the compositionality of T2I generation.
For example, GLIGEN \cite{li2023gligen} introduces trainable gated self-attention layers to integrate spatial inputs based on the pre-trained stable diffusion models.
ReCo \cite{yang2023reco} utilizes additional sets of positional tokens for T2I generation to achieve effective region control and fine-tune the pre-trained T2I models.
Feedback-based methods \cite{huang2023t2i, sun2023dreamsync, xu2024imagereward, fang2023boosting, lee2023aligning, lee2025parrot} use the generated images as feedback to optimize the T2I generation.
For instance, DreamSync \cite{sun2023dreamsync} utilizes a visual question answer model and an aesthetic quality evaluation model to recognize fine-grained discrepancies between the generated image and the textual input, thus enhancing the semantic alignment capabilities of the T2I model.
GORS \cite{huang2023t2i} fine-tunes pre-trained T2I models leveraging generated images aligned to the text prompt and text-image alignment reward-weighted loss.
Parrot \cite{lee2025parrot} jointly optimizes the T2I model with a multi-reward optimization strategy to improve the quality of image generation.
However, the above methods suffer from the following limitations: (i) lacking fine-grained and precise spatial information guidance, resulting in unrealistic spatial locations and relations in the generated images.
(ii) Difficulty in obtaining high-quality image feedback to effectively optimize the image generation.
(iii) Fine-tuning of the T2I model (\textit{e.g.,} stable diffusion) results in a large amount of computational and time overheads.

To address the above-mentioned problems, we propose a Multi-agent Collaboration-based Compositional Diffusion (MCCD) for high-quality text-to-image generation and complex scene generation.
Our novelty stems from three core contributions:
\textbf{(i)} We propose a multi-agent collaboration-based scene parsing module that constructs multiple agents with different tasks to implement collaboration in forward thought chain reasoning and backward feedback processes to precisely parse key scene elements.
\textbf{(ii)} Furthermore, Hierarchical Compositional diffusion is proposed to achieve sufficient interaction among parsed multiple scene elements and accurately generate complex scenes that match the text prompt.
\textbf{(iii)}  Comprehensive qualitative and quantitative experiments demonstrate that our method significantly improves the performance of the baseline models in a training-free manner, which has large advantages.

\section{Related Work}
\subsection{Text-to-Image Generation}
Text-to-Image (T2I) generation, \emph{i.e.}, text-conditional image synthesis, has been a key research hotspot in the field of multimodal learning \cite{ramesh2022hierarchical, betker2023improving}.
Numerous efforts have been devoted to generating visually natural and realistic images.
Generative Adversarial Networks (GANs) are typical T2I models that utilize adversarial training between the generator and the discriminator to produce images that are as close as possible to the real images.
In recent years, inspired by the application of Auto-Regressive Models (ARMs) in the field of text generation, many works have achieved favorable results in the field of T2I generation utilizing ARMs, such as  CogView\cite{ding2021cogview} and DALL-E 2/3 \cite{ramesh2022hierarchical, betker2023improving}.
Despite the progress achieved by the above studies, they still have many limitations, such as unstable training, difficult convergence, and unidirectional bias, which lead to poorer quality of image generation and lower generalizability.
Due to the natural fit of the inductive bias to the image data, diffusion models \cite{ho2020denoising,dhariwal2021diffusion, yang2023diffusion, sohl2015deep, rombach2022high}  are now widely used for T2I generation and significantly improve image generation quality and fidelity.
GLIDE \cite{nichol2021glide} utilizes the pre-trained CLIP model \cite{radford2021learning} to achieve semantic alignment between the text prompts and the generated images during the image sampling process.
Recent advances in T2I diffusion models have significantly improved the quality and realism of image generation in recent years, such as SDXL \cite{podell2023sdxl}, DALL-E 2/3 \cite{ramesh2022hierarchical, betker2023improving} and ContextDiff \cite{yang2024cross}.
In recent years, Large Language Models (LLMs) have been widely used in many tasks due to their powerful comprehension and reasoning capabilities \cite{li2023blip, yang2023baichuan, taylor2022galactica, zhu2023minigpt, chen2023llava, chowdhery2023palm}.
Many studies use Multimodal Large Language Models (MLLMs) in text-to-image (T2I) generation tasks and achieve performance gains \cite{yang2024mastering, qu2023layoutllm, hu2024ella, feng2024layoutgpt, liu2024llm4gen, zhong2023adapter, gani2023llm}. 
For example, RPG \cite{yang2024mastering} utilizes the Chain-of-Thought (CoT)  of MLLMs to extract layouts from text prompt to enhance T2I generation.
LMD \cite{lian2023llm} utilizes MLLMs to enhance the compositional generation of diffusion models by generating images grounded on bounding box layouts from the MLLMs.
However, the above methods have the following limitations:(1) Simply using MLLMs to process text prompts without sufficiently exploiting and utilizing their powerful comprehension and inference capabilities. (2) Only utilizing MLLMs as a layout generator to control image synthesis, neglecting the extraction of other important scene elements.
To address the above problems, we propose a Multi-agent Collaboration-based scene Parsing (MCP) module, which constructs a multi-agent system consisting of multiple agents with various divisions of labor, and utilizes multi-agent collaboration and interaction to achieve adequate extraction and parsing of key scene elements in text prompt, thereby facilitating the subsequent scene generation process.

\subsection{Compositional Diffusion Generation}
In recent years, many methods have been introduced to improve compositional T2I generation \cite{yang2023reco, li2023gligen, mou2024t2i, lian2023llm, zhang2023adding} to enhance the capabilities of diffusion models in terms of attribute binding, object relations, and numeracy. For example, ReCo \cite{yang2023reco} and GLIGEN \cite{li2023gligen} introduce a location-aware adapter in diffusion models to improve the spatial plausibility of the generated images. 
LMD \cite{lian2023llm} utilizes LLM to generate scene layouts and designs a controller to bootstrap the pre-trained diffusion model. 
RPG \cite{yang2024mastering} denoises each subregion in parallel and applies a post-processing step of resizing and concatenation for high-quality compositional generation. 
The T2I-Adapter \cite{mou2024t2i} facilitates compositional T2I generation by controlling the semantic structure through some high-level features of the image.
However, these methods can only implement coarse compositional control, leading to unsatisfactory image generation results, especially when dealing with complex prompts.
Therefore, we propose hierarchical compositional diffusion, which utilizes more precise control to progressively refine image synthesis.

\begin{figure*}[t]
\centering
\includegraphics[width=1.0\textwidth]{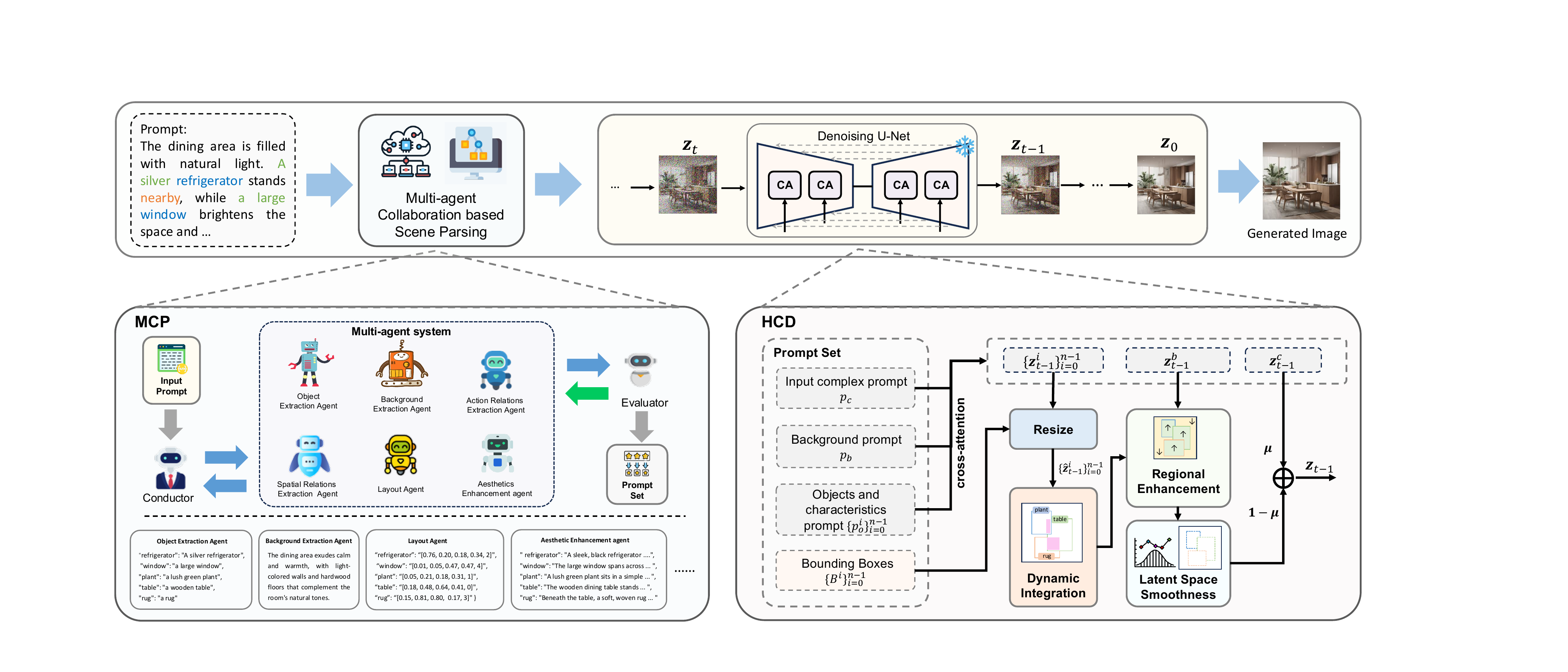}
\caption{\textbf{The overall framework of the proposed MCCD}. MCCD consists of two core components: Multi-agent Collaboration-based scene Parsing (MCP) module and Hierarchical Compositional Diffusion (HCD) module. In MCP, the blue and green arrows indicate forward CoT reasoning and backward  feedback processes, respectively}
\label{fig:framework}
\end{figure*}

\section{Methodology}
\subsection{Overall Framework}
As shown in \Cref{fig:framework}, given a complex text prompt containing multiple objects and relations, the goal of MCCD is to produce realistic and high-quality images. Our MCCD is a training-free framework with the following workflow: 
(i) Multi-agent collaboration-based scene parsing module utilizes a multi-agent collaborative approach to parse individual elements in a text prompt. (2) Hierarchical compositional diffusion is used to interact with scene elements and achieve high-quality image generation using dynamic integration, regional enhancement, and latent space smoothing.

\subsection{Multi-agent Collaboration based Scene Parsing}
Based on the sufficient consideration of the composition of complex scenes, we categorize the scene elements into several parts: objects and their characteristics, backgrounds, action relations, spatial relations, and layouts.
To generate complex scenes, we need to explore all the key elements in the text prompt thoroughly.
The previous method \cite{yang2024mastering} utilizes the Chain-of-Thought (CoT) capability of MLLMs to extract objects and layouts from the input text prompt to a certain extent. 
Despite some success, when dealing with more complex scenarios, the CoT-based approach often fails to sufficiently parse the intricate relations in the scene. Furthermore, unidirectional CoT reasoning lacks error correction mechanisms, leading to inaccurate elemental parsing and image generation.
Therefore, we propose a Multi-agent Collaboration-based Scene Parsing (MCP) module that splits the scene parsing process into multiple sub-stages and constructs corresponding specialized agents for each stage, and generates a multi-agent system simultaneously. 
In the system, each agent dynamically collaborates and interacts with other agents while strictly performing its specialized tasks, thus maximizing the advantages of teamwork.
This paradigm ensures that MLLMs accurately parse multiple elements contained in text prompts, providing significant advantages over traditional CoT reasoning.

The constructed multi-agent system consists of six agents whose definitions and tasks are: 
(1) object extraction agent, whose task is to extract objects and their characteristics from text prompts. (2) Background extraction agent, whose task is to extract the object-independent background descriptions. (3) Action relations extraction agent, whose task is to capture the action relations between the objects. (4) Spatial relations extraction agent, whose task is to capture the spatial relations between the objects. (5) Layout agent, whose task is to conceptualize the layout of objects for a reasonable composition. (6) Aesthetics enhancement agent, whose task is to beautify the objects' characteristics to enhance the image aesthetics. 
Our goal is to integrate the outputs of multiple agents to ultimately generate the objects and characteristics prompt, background prompt, and bounding boxes of all objects.
Each agent is defined by an MLLM and is associated with a prompt template while allowing access to an external knowledge base.

Moreover, we construct a conductor $\mathcal{C}$ and an evaluator $\mathcal{E}$ to coordinate multiple agents for effective collaboration. Specifically, the conductor dynamically directs different agents to construct a forward CoT in iterations and passes candidate answers to the external program execution environment. The evaluator then uses feedback signals to trigger a backward feedback process. These two processes are elaborated below.

\noindent \textbf{Forward CoT Reasoning.}
In forward CoT reasoning, the conductor $\mathcal{C}$ adaptively determines the entire set of participating agents, casting the procedure as a sequential decision-making problem in which each possible action corresponds to a particular agent selection.
We define the input prompt as $\mathcal{P}$ and a set of predefined agents as $\xi =\{\mathcal{A}_{\phi_1}, \mathcal{A}_{\phi_2}, \dots, \mathcal{A}_{\phi_n}\}$, where $n$ is the total number of agents and $\phi_i$ is the configuration of the agent $i$-th.
The set of output for the $t$-th inference step is denoted as $\mathcal{O}_t$, and the state is denoted as $\mathcal{S}_t = \left(\mathcal{P}, \mathcal{O}_t, t\right)$.
With the powerful prompt learning capability of MLLM, agents can achieve the same functionality in a training-free manner as decision-making agents that require large amounts of data for training in traditional reinforcement learning, with significant advantages in terms of efficiency and overhead.
Thus, we take the conductor as a policy function for selecting an agent, denoted as:
\begin{equation}
    \mathcal{C}(a \mid s)=P_r\left\{\mathcal{A}_{\phi_t}=a \mid S_t=s\right\}.
\end{equation}
The agent selection strategy can be translated into the design of the prompt template, which requires prompt engineering to realize the optimal strategy. 
Each step of forward CoT reasoning is denoted as:
\begin{equation}
\mathcal{A}_{\phi_{i_t}}=\mathcal{C}\left(S_t\right),
\end{equation}
\begin{equation}
o=\mathcal{A}_{\phi_{i_t}}\left(\mathcal{P}, \mathcal{O}_t\right),
\end{equation}
\begin{equation}
\mathcal{O}_{t+1}=\mathcal{O}_t \cup\{o\},
\end{equation}
where $\mathcal{A}_{\phi_{i_t}}$ represents the selected $i_t$-th agent at step $t$ and $o$ denotes the output of the selected agent.
After reaching the maximum step $T$, the forward process is aborted and all outputs are integrated into the result $\mathcal{R}$.

\noindent\textbf{Backward feedback.}
The backward feedback strategy enables a multi-agent system to adjust collaborative behavior by using the evaluator's evaluation of the forward response.
The execution order of the agents is defined as $\eta = \{\mathcal{A}_{\phi_{i_1}}, \mathcal{A}_{\phi_{i_2}}, \dots, \mathcal{A}_{\phi_{i_n}} \}$, where $i_t$ denotes the index of the agent at step $t$.
The backward feedback process begins with an external feedback $f_{raw}$, usually provided by the program execution environment, denoted as 
\begin{equation}
    f_{raw} = execution(\mathcal{R}).
\end{equation}
The raw feedback is then evaluated by evaluator $\mathcal{E}$ to obtain an initial signal: $(f_0, sf_0) = \mathcal{E}(f_{raw})$, where $f_0$ indicates whether the backward process needs to continue or not, and $sf_0$ indicates the localization of the error in the backward feedback.
If the result $\mathcal{R}$ in the forward reasoning is correct, then $f_0$ is set to false and the whole reasoning process ends. Otherwise, the conductor $\mathcal{C}$ initiates a backward feedback process that updates the response by back-propagating from the last agent.
At step $t$ backward, the state update is represented as follows:
\begin{equation}
\left(f_t, s f_t\right) \leftarrow feedback \left(\mathcal{A}_{\phi_{i_{T-t+1}}}, \mathcal{P}, \mathcal{O}_t, f_{t-1}\right),
\end{equation}

\begin{equation}
\mathcal{O}_{t+1}=\mathcal{O}_t \cup\left\{sf_t\right\}.
\end{equation}
The backward feedback process continues to be executed iteratively until the feedback signal indicates that the agent has made a mistake or until all the agents have undergone reflection. Once this occurs, the forward process will then be executed again. This process is repeated continuously until a reasonable answer is produced.

\begin{figure*}[t]
  \centering
  \includegraphics[width=\textwidth]{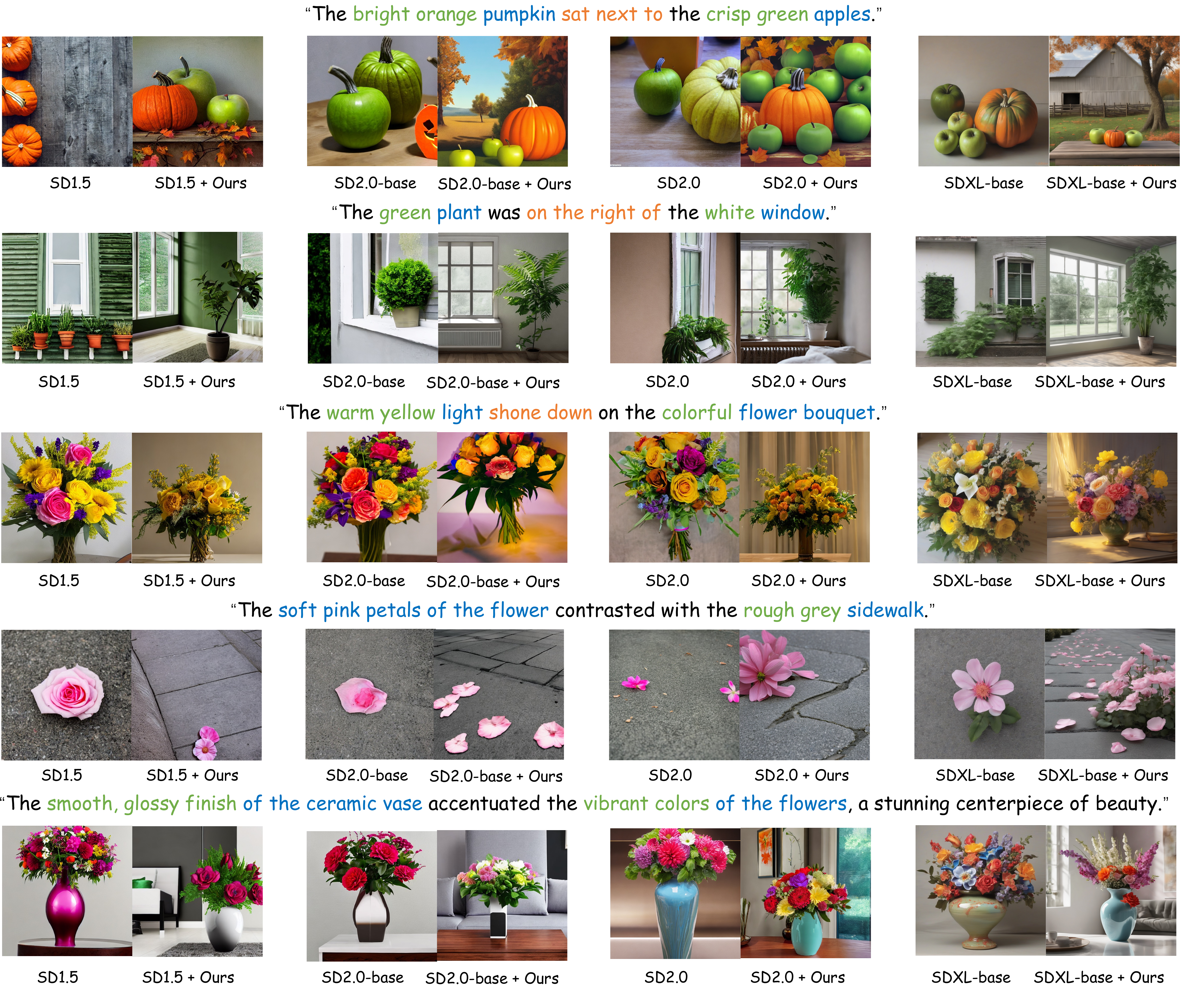}
  \caption{\textbf{Qualitative results of MCCD improving diffusion models}. MCCD enhances the attribute binding and spatial relationships of the base diffusion models. The generated results have reasonable backgrounds and detailed textures with great aesthetics and realism.}
  \label{fig:cmp}
\end{figure*}

\begin{table*}[t]
\centering
\caption{\textbf{Evaluation results on T2I-CompBench}. MCCD shows the best performance in terms of Attribute Binding, Object Relationship, and Complex. Basic data is derived from \cite{huang2023t2i}.}
\renewcommand{\arraystretch}{1.5}
\setlength{\tabcolsep}{16pt}
\label{tab:my-table}
\resizebox{\textwidth}{!}{%
\begin{tabular}{ccccccc}
\toprule
\multirow{2}{*}{Models}                 & \multicolumn{3}{c}{Attribute    Binding}            & \multicolumn{2}{c}{Object    Relationship} & \multirow{2}{*}{Complex↑} \\ \cmidrule{2-6}
                                        & Color↑          & Shape↑          & Texture↑        & Spatial↑             & Non-Spatial↑        &                           \\ \midrule
Composable Diffusion \cite{liu2022compositional}  & 0.4063          & 0.3299          & 0.3645          & 0.0800               & 0.2980              & 0.2898                    \\
Structured Diffusion \cite{feng2022training} & 0.4990          & 0.4218          & 0.4900          & 0.1386               & 0.3111              & 0.3355                    \\
Attn-Exct v2 \cite{chefer2023attend}       & 0.6400          & 0.4517          & 0.5963          & 0.1455               & 0.3109              & 0.3401                    \\
GORS \cite{huang2023t2i}             & 0.6603          & 0.4785          & 0.6287          & 0.1815               & 0.3193              & 0.3328                    \\
DALL-E 2 \cite{ramesh2022hierarchical}           & 0.5750          & 0.5464          & 0.6374          & 0.1283               & 0.3043              & 0.3696                    \\
PixArt-$\alpha$ \cite{chen2023pixart}           & 0.6886          & 0.5582          & 0.7044          & 0.2082               & 0.3179              & 0.4117                    \\ 
\midrule
SD1.5 \cite{rombach2022high}            & {0.3134} & {0.2954} & {0.3772} & {0.1087} & {0.3015} & {0.2994} \\
SD1.5 + \textbf{MCCD}      & \textbf{0.3508} & \textbf{0.3193} & \textbf{0.4026} & \textbf{0.1462} & \textbf{0.3078} & \textbf{0.3054} \\
\midrule
SD2.0-base \cite{rombach2022high}       & {0.4498} & {0.3584} & {0.4319} & {0.1140} & {0.3045} & {0.3021} \\
SD2.0-base + \textbf{MCCD} & \textbf{0.4823}                        & \textbf{0.3702}                        & \textbf{0.4621}                        & \textbf{0.1613}                        & \textbf{0.3106}                        & \textbf{0.3146}                        \\
\midrule
SD2.0 \cite{rombach2022high}            & 0.4852                        & 0.4066                        & 0.4591                        & 0.1490                        & 0.2905                        & 0.3173                        \\
SD2.0 + \textbf{MCCD}      & \textbf{0.5090}                        & \textbf{0.4170}                        & \textbf{0.4921}                        & \textbf{0.1785}                        & \textbf{0.3123}                        & \textbf{0.3250}                        \\
\midrule
SDXL-base \cite{betker2023improving}        & {0.5744} & {0.4705} & {0.4907} & {0.1971} & {0.3009} & {0.3130} \\
SDXL-base + \textbf{MCCD}  & \textbf{0.6278}                        & \textbf{0.4832}                        & \textbf{0.5647}                        & \textbf{0.2350}                        & \textbf{0.3132}                        & \textbf{0.3348}       \\                
 \bottomrule
\end{tabular}%
}
\end{table*}

\subsection{Hierarchical Compositional Diffusion}
Previous region-based diffusion methods, such as \cite{yang2024mastering}, divide images into multiple complementary regions. It takes into account spatial relations to a certain extent, but it cannot cope with more complex scenes due to the non-overlapping and independent properties of regions.
Additionally, \cite{xie2023boxdiff} imposes constraints on cross-attention maps so that the positions and sizes of objects are as consistent as possible with the bounding box, but ignores the consideration of the overlaps between the bounding boxes and smoothness near the bounding box, leading to unrealistic scene synthesis.
To this end, we propose a Hierarchical Compositional Diffusion (HCD) module that performs progressive interaction of multiple elements obtained from scene parsing. Specifically, we dynamically balance the overlapping regions between multiple objects using the Gaussian mask and a regional enhancement strategy is used to enlarge the discrepancy between objects and background in the latent space. Moreover, Gaussian smoothing is used to enhance the smoothness around the bounding box.

As shown in \Cref{fig:framework}, the MCP parses a complex prompt with n objects into multiple prompts, \emph{i.e.}, the input complex prompt $p_c$, the object prompts $\{p_o^i\}^{n-1}_{i=0}$ and background prompt $p_b$. 
At each timestep, we feed each prompt into the denoising network in parallel, using cross-attention layers to generate the corresponding latent representations.
\begin{equation}
\small
    z_{t-1}=Softmax\left(\frac{\left(W_Q \cdot \phi\left(z_t\right)\left(W_K \cdot \psi\left(p\right)\right)\right.}{\sqrt{d}}\right)\left(W_V \cdot \psi\left(p\right)\right),
\end{equation}
where image latent $\bm{z}_t$ is the query and each prompt is the key and the value. $\bm{W}_Q$,$\bm{W}_K$,$\bm{W}_V$ are linear projections and $d$ is the latent projection dimension of the keys and queries.
The latent representations of complex prompt, all object prompts, and background prompt are denoted as $z_{t-1}^c$, $\{z_{t-1}^i\}_{i=0}^{n-1}$, and $z_{t-1}^b$.
We use bilinear interpolation to resize the latent representation of the object based on the dimensions of the bounding box, denoted as follows:
\begin{equation}
    \hat{\bm{z}}_{t-1}^i = R \left(z_{t-1}^i, B^i\right).
\end{equation}

Complex scenes often contain multiple objects with intricate positions and action relationships across multiple objects, resulting in a large number of overlapping regions. To this end, we design a dynamic integration mechanism based on a depth-aware Gaussian mask to achieve adaptive and smooth feature fusion in overlapping regions.
In general, the center region of the bounding box contains the core features of the object, while the features in the edge region gradually decrease in importance. To keep as many core features as possible in the overlapping regions of the bounding box, for each bounding box $(x_0,y_0,w,h)$, we construct a Gaussian mask matrix that contains weights that gradually decay from the center to the edges, represented as follows:
\begin{equation}
    M(x, y)=\exp \left(-\frac{\left(x-\mu_x\right)^2+\left(y-\mu_y\right)^2}{2 \sigma^2}\right),
\end{equation}
where $\left(\mu_x, \mu_y\right)=(x_0 + w / 2, y_0 + h / 2)$ is the center of the mask matrix, and the $\sigma=max(w,h) / 2$ is the standard deviation controlling the width of the Gaussian distribution.
Furthermore, an object's layer depth $d$ indicates its front and back positions in the image. Objects with smaller layer depth appear closer in the image, so greater weights should be assigned during feature fusion.
To achieve a smooth transition between foreground and background, we calculate continuous and eased layer depths:
\begin{equation}
    w_i=\frac{1}{1+\exp \left(\alpha \cdot\left(d_i - \frac{n-1}{2} \right)\right)},
\end{equation}
where $d \in \{0,1,\dots, n-1\}$ is the layer depth of the object, and $\alpha$ is the smoothness control parameter.
Given any coordinates $(x,y)$ of the overlapping regions of the bounding boxes, their dynamically fused features are represented as.
\begin{equation}
    \hat{\bm{z}}_{t-1}^i(x, y)=\frac{\sum_{i=1}^{m} w_i \cdot M_i(x, y) \cdot z^i_{t-1}(x, y)}{\sum_{i=1}^{m} w_i \cdot M_i(x, y)},
\end{equation}
where $m$ is the number of overlapping regions. 
Then, we concatenate the latent representations of all object prompts based on the positions of the bounding boxes to achieve control over positional relations. The area not covered by the bounding box is filled by the latent representation of the background. This process is represented as:
\begin{equation}
\small
    \bm{z}^\prime_{t-1} = Concat(\{\bm{M}_{B^i} \cdot \hat{\bm{z}}^i_{t-1} \}^{n-1}_{i=0}, \{\bm{M}_{\neg \left( \bigcup_{i=0}^{n-1} B^i \right)} \cdot \bm{z}_{t-1}^b\}),
\end{equation}
where $\bm{M}_{B^i}$ and $\bm{M}_{\neg \left( \bigcup_{i=0}^{n-1} B^i \right)}$ are masks denoting the region within the bounding box $B^i$ and the background region outside all bounding boxes, respectively.

To ensure that objects are generated within the designated bounding box regions, we implement a regional enhancement mechanism that emphasizes the latent representation of the bounding box regions and simultaneously suppresses the influence of surrounding background areas.
\begin{equation}
\bm{z}^\prime_{t-1}+=\lambda_{\text {pos }} \cdot \bm{M}_{B^i} \cdot\left(\operatorname{Max}\left(\hat{z}_{t-1}^i\right)-\bm{z}^\prime_{t-1}\right),
\end{equation}
\begin{equation}
    \bm{z}^\prime_{t-1}-=\lambda_{\text {neg }} \cdot\left(\bm{M}_{\neg \left( \bigcup_{i=0}^{n-1} B^i \right)}\right) \cdot\left(\bm{z}^\prime_{t-1}-\operatorname{Min}\left(\hat{z}_{t-1}^i\right)\right),
\end{equation}
where $\lambda_{\text {pos }}$ and $\lambda_{\text {neg }}$ are both set to 0.2.

The features inside and outside the bounding box have large discrepancies, resulting in an unsmooth excess of the generated image near the bounding box. Therefore, we employ Gaussian filtering to smooth the features near the bounding box in the latent space. 
First, we define a two-dimensional Gaussian kernel, denoted as follows:
\begin{equation}
    G_\sigma(i, j)=\frac{1}{2 \pi \sigma^2} \exp \left(-\frac{i^2+j^2}{2 \sigma^2}\right),
\end{equation}
where $\sigma=1.0$ is the standard deviation that controls the spread of the Gaussian filtering.
Then, assuming an arbitrary feature with coordinates $(x,y)$ near the bounding box, we perform Gaussian filtering on it, denoted as follows:
\begin{equation}
    z_{t-1}^{smooth}(x, y)=\sum_{i=-k}^k \sum_{j=-k}^k z^\prime_{t-1}(x+i, y+j) \cdot G_\sigma(i, j).
\end{equation}
Furthermore, to preserve the overall semantics and distribution of the image as much as possible, the weighted sum of $z^c_{t-1}$ and $z_{t-1}^{smooth}$ is used to produce the final denoised output that maintains consistency between image and text.
\begin{equation}
    z_{t-1}=\mu \cdot z_{t-1}^c+(1-\mu) \cdot z_{t-1}^{smooth},
\end{equation}
where $\mu$ is the trade-off weight set to 0.8.

\begin{figure*}[t]
\centering
\includegraphics[width=0.9\textwidth]{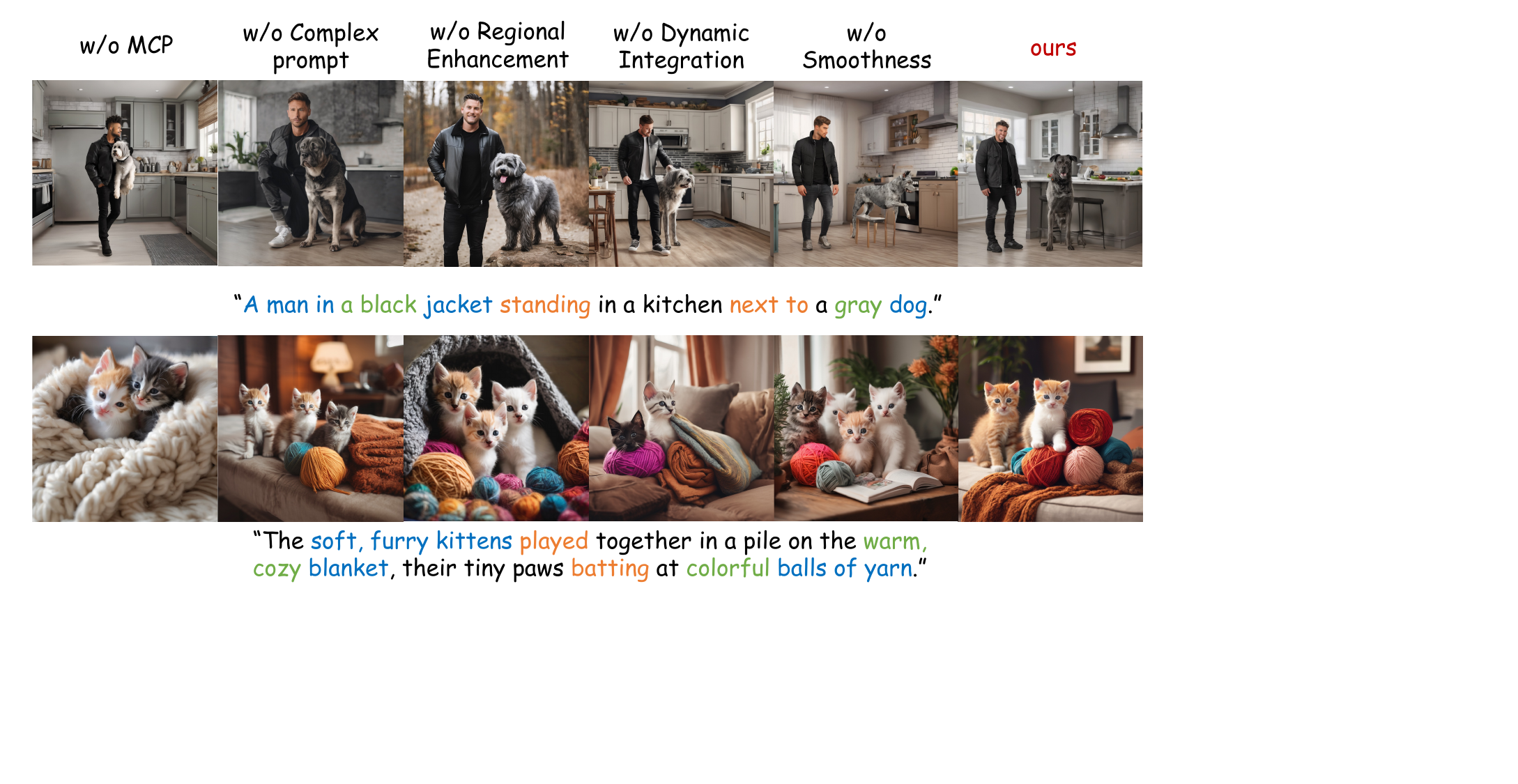}
\caption{\textbf{Ablation results of MCCD}. The poor results after removing the critical components prove that each component is crucial.}
\label{fig:ablation}
\end{figure*}

\begin{figure*}[t]
\centering
\includegraphics[width=1.0\textwidth]{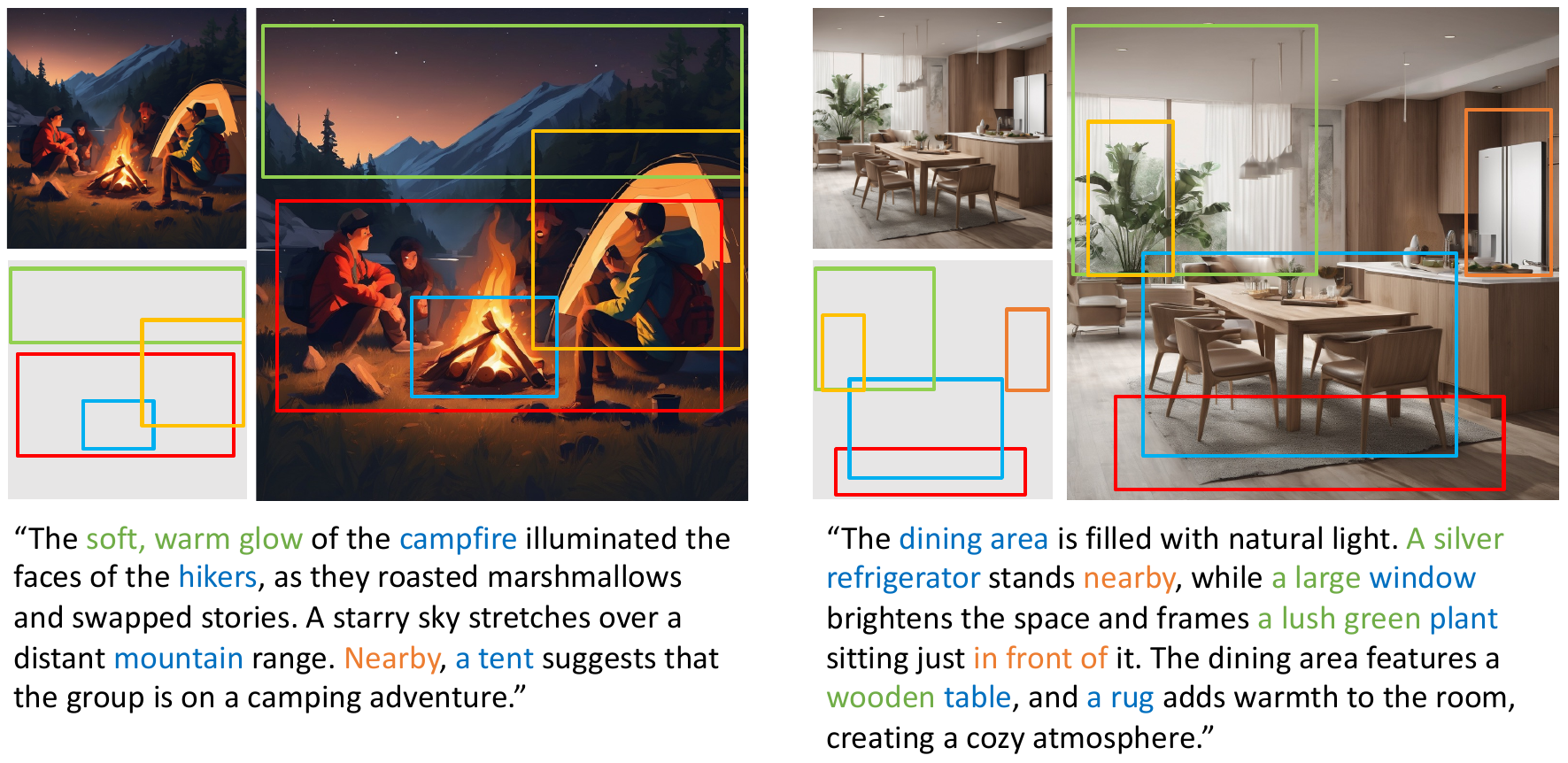}
\caption{\textbf{Additional qualitative results}. MCCD can handle complex text prompts with multiple objects and attribute binding relationships effectively, generating reasonable bounding box layouts and producing aesthetically pleasing images with high realism.}
\label{fig:vis}
\end{figure*}

\section{Experiments}
\subsection{Datasets and Evaluation Metrics}
Since our framework is training-free, no dataset is required for training. 
To comprehensively evaluate the performance of different methods, we selected six metrics from three categories in the T2I-CompBench benchmark \cite{huang2023t2i}, namely Attribute Binding (Color, Shape, Texture), Object Relationship (Spatial, Non-Spatial), and Complex.
The three metrics of Attribute Binding are evaluated by disentangled BLIP-VQA. The spatial metric is determined by detecting the object through UniDet and comparing the center of the object's bounding box in the image to determine the spatial relationship. CLIP-Score is applied for the evaluation of the Non-Spatial metric. The Complex metric is the average score of disentangled BLIP-VQA, CLIP-Score, and UniDet. 
Each metric corresponds to 300 prompts, and each prompt yields 10 images for evaluation with diverse seeds.

\subsection{Implementation Details}
Our MCCD is both generic and extensible, allowing us to seamlessly integrate a wide range of MLLM architectures and diffusion models into the framework. 
In our experiments, we use GPT-4o-mini \cite{achiam2023gpt} to construct multiple agents in MCP.
The pre-trained diffusion models in HCD consist of: SDv1.5 \cite{rombach2022high}, SDv2 \cite{rombach2022high}, SDv2-base \cite{rombach2022high}, and SDXL-base \cite{podell2023sdxl}.
The inference process is set to 20 steps, with the classifier-free guidance scale set to 7.0. We have carefully designed task templates and selected high-quality context examples to ensure optimal performance. 
All experiments are conducted on a single A800 GPU. 

\subsection{Main Results}
\noindent \textbf{Qualitative Results.}
 \Cref{fig:cmp} shows the results of MCCD on multiple diffusion models. We have the following observations
({\romannumeral1}) Attribute Binding: Attributes (\emph{e.g.}, color, quantity) are incorrectly bound to objects in the generated results of the base diffusion models, while each attribute is correctly bound to the corresponding object after applying MCCD. 
({\romannumeral2}) Spatial location relationships are often incorrect in the results generated from the base diffusion models. For example, objects may be incorrectly placed in unnatural locations, or spatial relationships between objects may be confused. The spatial relationship confusion problem is solved after applying MCCD.
({\romannumeral3}) Realism and aesthetics: Many of the generated results of the base diffusion models lack suitable backgrounds and detailed texture structures, while MCP provides detailed, realistic, and aesthetically pleasing descriptions of objects and backgrounds for the complex prompts, which makes the results highly aesthetic.

\Cref{fig:vis} shows the results of complex prompts with multiple objects and attribute binding by MCCD. Through in-depth understanding and parsing of the input complex prompts, MCCD can effectively recognize and extract the objects and their interrelationships in the text prompt and then generate a reasonable and accurate bounding-box layout. The position and size of each bounding box closely correspond to the elements (e.g., objects, colors, shapes, etc.) in the complex prompt, ensuring the accurate positioning and realistic rendering of each element in the image. 
In addition, MCCD fully captures the connection between the object and its surroundings, resulting in an image that exhibits a high degree of realism and aesthetics.
In this way, MCCD can transform complex and layered text prompts into clearly structured and highly realistic images, perfectly combining text and visual expression.

\noindent \textbf{Quantitative Results.}
\Cref{tab:my-table} shows the results of multiple open source reproducible models and MCCD in T2I-CompBench. We can draw the following conclusions: 
({\romannumeral1}) Overall, the best results are achieved by applying MCCD to SDXL, with an overall metric improvement of 9.04\%, which is significantly better than the baseline models. 
({\romannumeral2}) Compared to other diffusion models, the application of MCCD resulted in the highest overall improvement in metrics of 9.04\%, demonstrating the general scalability of MCCD. 
({\romannumeral3}) Benefiting from MCP, the key components of MCCD, the Spatial metric of the diffusion model is increased by up to 41.49\%, and the Complex metric is increased by up to 8.73\%, which effectively improves the spatial position relationship, semantic expressiveness, and image fidelity of the generated images.

\subsection{Ablation Study}
To verify the necessity of each component, we conduct full-scale ablation studies. The results are shown in \Cref{fig:ablation}. 
({\romannumeral1}) Firstly, MCP is removed from MCCD. The poor attribute binding and spatial location relationship indicate that it is crucial to dynamically invoke multiple agents to generate reasonable bounding boxes, objects, and background descriptions before image generation. 
({\romannumeral2}) In addition, we remove the complex prompt in HCD, and the image generation results that are inconsistent with the prompt descriptions suggest that the complex prompt is indispensable for maintaining consistency between image and text.
({\romannumeral3}) Then, Regional Enhancement is removed from HCD. The phenomenon of poor semantic representation of the objects and the background in the image suggests that regional enhancement is crucial. 
({\romannumeral4}) Additionally, we remove Dynamic Integration, and the image appears to be partially missing in the occluded object's part, presenting an anomalous structure. 
({\romannumeral5}) Finally, we remove Latent Space Smoothness, and the image exhibits unnatural transitions where the bounding box connects to the background, indicating that smoothness is essential for natural transitions between the bounding boxes and between the bounding box and the background in the image.

\section{Conclusion}
In this paper, we propose Multi-Agent Collaboration-based Compositional Diffusion (MCCD) for generating high-quality complex scenes. Specifically, we design a multi-agent collaboration-based scene parsing module to fully extract the scene elements contained in the text by constructing a multi-intelligentsia system based on MLLMs. In addition, we propose a hierarchical compositional diffusion that utilizes dynamic integration of overlapping regions, regional enhancement, and latent space smoothness to generate realistic and aesthetically pleasing images.
Comprehensive experiments prove the advantages of our method.

\newpage



\maketitlesupplementary

\section{The Prompt design of MCP}
We provide detailed descriptions and prompt template implementations for the conductor, evaluator, and all agents. The text enclosed within the curly braces denotes placeholders that will be dynamically populated during runtime based on the input text prompt and the agent's output.
\subsection{Conductor}

The role of a conductor is highly specialized and significant, which necessitates a more intricate prompt design compared to other agents. 
The task of the conductor is to coordinate all the agents you manage so that they can work together to solve the problem.
The prompt template for a conductor is shown in \Cref{tb: Conductor}.

\begin{table}[t]
    \centering
    \caption{The prompt template for the conductor.}
    \label{tb: Conductor}
    \begin{tabular}{@{}p{0.95\linewidth}@{}} 
        \toprule
        \textit{Conductor} \\ 
        \midrule
        \texttt{[INST]<SYS>} \\
        You are the leader of an agent system for text parsing in complex scenes. Now, you need to coordinate all the agents you manage so that they can work together to solve the problem. Next, you are given a specific text prompt, and your goal is to select the agents you think are best suited to solicit insights and suggestions.
        Generally speaking, the parsing of complex scenes includes several processes: object extraction, background extraction, relation extraction, layout extraction, and aesthetic optimization. Different text prompts may correspond to different processes, so you need to select the corresponding agent to solve the problem dynamically.
        \texttt{</SYS>} \\
        \midrule
        \texttt{<USER>} \\
        The text prompt is: \texttt
        {\{text prompt\}}.
        
        Remember, based on the capabilities of different agents and the current status of the problem-solving process, you need to decide which agent to consult next. The agents' capabilities are described as follows: \texttt{\{agent info\}}. 
        
        Agents that have already outputted their answers include: \texttt{\{outputted agents\}} .

        Please select an agent to consult from the remaining agents \texttt{\{remaining agents\}}.
        
        Remember, the agent must choose from the existing list above.
        
        Note that you must complete the workflow within the remaining \texttt{\{remaining steps}\} steps.

        You should output the name of the agent directly. The next agent is:
        \texttt{</USER>[INST]}
        \\
        \bottomrule
    \end{tabular}
    \vspace{-6pt}
\end{table}

\subsection{Evaluator}
The evaluator's task is to exercise critical thinking to assess the truthfulness and reasonableness of the agent's output. If it is not reasonable, then make recommendations for modification.
The prompt template for the evaluator is given in \Cref{tb: Evaluator}.

\begin{table}[t]
    \centering
    \caption{The prompt template for the evaluator.}
    \label{tb: Evaluator}
    \begin{tabular}{@{}p{0.95\linewidth}@{}} 
        \toprule
        \textit{Evaluator} \\ 
        \midrule
        \texttt{[INST]<SYS>} \\
        Your task is to exercise critical thinking to assess the truthfulness and reasonableness of the agent's output. If it is not reasonable, then make recommendations for modification.

        Output format: \{``Result'':``Evaluation results, with a value of right or wrong'', ``Problem'': ``If there is a problem, describe it in detail, otherwise, the value is null'', ``Modification Suggestion'': ``If the result is incorrect, describe the proposed change, otherwise, the value is null''\}.
        \texttt{</SYS>} \\
        \midrule
        \texttt{<USER>} \\
        The input prompt is described as \texttt{\{input prompt\}}.
        
        The output of all agents is \texttt{\{outputs\}}.

        Please evaluate the reasonableness of the agents' outputs. If they are not reasonable, please state your suggestions for modification.
        \\
        \texttt{</USER>[INST]}
        \\
        \bottomrule
    \end{tabular}
\end{table}

\subsection{Agent System}
In this section, we provide an in-depth overview of the individual agents involved in our MCCD. 
Each agent is assigned a specific role and domain knowledge related to problem-solving.

\noindent \textbf{Object extraction agent.}
The object extraction agent's task is to extract key entities and their corresponding characteristics from the text input prompt. 
The prompt template for the object extraction agent is shown in \Cref{tb:object extraction agent}.

\begin{table}[t]
    \centering
    \caption{The prompt template for the object extraction agent.}
    \label{tb:object extraction agent}
    \begin{tabular}{@{}p{0.95\linewidth}@{}} 
        \toprule
        \textit{Object extraction agent} \\ 
        \midrule
        \texttt{[INST]<SYS>} \\
        As an object extraction agent, you extract key entities and their corresponding characteristics from the text input prompt. 
        
        Extract multiple object and characteristic pairs if multiple characteristics of an entity describe different parts of a person, such as the head, clothes/body, and underwear.
        
        To ensure numeric accuracy, objects with the same class name (\emph{e.g.}, five apples) will be separately assigned to different regions.

        The output format is  \{object$_1$:characteristic$_1$, object$_2$:characteristic$_2$, \dots, object$_n$:characteristic$_n$\}
        
        \texttt{</SYS>} \\
        \midrule
        \texttt{<USER>} \\
        The input prompt is described as: \texttt{\{input prompt\}}.

        You are supposed to refer to the output of other agents: \texttt{\{outputs\}}.

        Please output the extracted objects and their characteristics in the text prompt.
        \\
        \texttt{</USER>[INST]}
        \\
        \bottomrule
    \end{tabular}
\end{table}

\noindent \textbf{Background extraction agent.}
The task of the background extraction agent is to extract the background from this complex prompt. 
The extracted background is required not to contain any object and its characteristics, but only a description of the scene as a whole.
The prompt template for the background extraction agent is shown in \Cref{tb: background extraction agent}.

\begin{table}[t]
    \centering
    \caption{The prompt template for the background extraction agent.}
    \label{tb: background extraction agent}
    \begin{tabular}{@{}p{0.95\linewidth}@{}} 
        \toprule
        \textit{Background extraction agent} \\ 
        \midrule
        \texttt{[INST]<SYS>} \\
        As an object extraction agent, your task is to extract the background from this complex prompt. The extracted background is required not to contain any object and its characteristics, but only a description of the scene as a whole.
        
        \texttt{</SYS>} \\
        \midrule
        \texttt{<USER>} \\
        The input prompt is described as: \texttt{\{input prompt\}}.
        
        The outputs given by other agents are as follows: \texttt{\{outputs\}}, please refer to them carefully.

        Please extract and output the background in the text prompt.
        \\
        \texttt{</USER>[INST]}
        \\
        \bottomrule
    \end{tabular}
\end{table}

\noindent \textbf{Action relations extraction agent.}
The task of the action relation extraction agent is to extract the action relations between objects in the scene, such as holding, sitting, and so on, to bind multiple objects.
Its prompt template is illustrated in \Cref{tb: Action relations extraction agent}.
\begin{table}[t]
    \centering
    \caption{The prompt template for the action relations extraction agent.}
    \label{tb: Action relations extraction agent}
    \begin{tabular}{@{}p{0.95\linewidth}@{}} 
        \toprule
        \textit{Action relation extraction agent} \\ 
        \midrule
        \texttt{[INST]<SYS>} \\
        As an action relation extraction agent, your task is to extract the action relations between objects in the scene, such as holding, sitting, and so on, to fully extract the spatial information of the scene.

        The output format is \{(object$_1$, action relation$_1$, object$_2$), \dots, ((object$_2$, action relation$_2$, object$_n$)\}
        
        \texttt{</SYS>} \\
        \midrule
        \texttt{<USER>} \\
        The input prompt is described as: \texttt{\{input prompt\}}.
        
        The outputs given by other agents are as follows: \texttt{\{outputs\}}.

        Please extract and output the action relations between objects in the text prompt.
        \\
        \texttt{</USER>[INST]}
        \\
        \bottomrule
    \end{tabular}
\end{table}

\noindent \textbf{Spatial relations extraction agent.}
The task of the spatial relations extraction agent is to extract the spatial relations between objects in the scene, such as left, beside, and so on, to fully extract the spatial information of the scene.
Its prompt template is illustrated in \Cref{tb: Spatial relations extraction agent}.
\begin{table}[h]
    \centering
    \caption{The prompt template for the spatial relations extraction agent.}
    \label{tb: Spatial relations extraction agent}
    \begin{tabular}{@{}p{0.95\linewidth}@{}} 
        \toprule
        \textit{Spatial relation extraction agent} \\ 
        \midrule
        \texttt{[INST]<SYS>} \\
        As a spatial relation extraction agent, your task is to extract the spatial relations between objects in the scene, such as ``left'' and ``beside'', to fully capture the spatial information of the scene.

        The output format is \{(object$_1$, spatial relation$_1$, object$_2$), \dots, (object$_2$, spatial relation$_2$, object$_n$)\}

        \texttt{</SYS>} \\
        \midrule
        \texttt{<USER>} \\
        The input prompt is described as: \texttt{\{input prompt\}}.
        
        You should refer to the output of other agents: \texttt{\{outputs\}}.

        Please extract and output the spatial relations between objects in the text prompt.
        \\
        \texttt{</USER>[INST]}
        \\
        \bottomrule
    \end{tabular}
\end{table}

\noindent \textbf{Layout agent.}
The task of the layout agent is to layout the scene by generating a bounding box for each object. 
Its prompt template is illustrated in \Cref{tb: Layout agent}.

\begin{table}[h]
    \centering
    \caption{The prompt template for the layout agent.}
    \label{tb: Layout agent}
    \begin{tabular}{@{}p{0.95\linewidth}@{}} 
        \toprule
        \textit{Layout agent} \\
        \midrule
        \texttt{[INST]<SYS>} \\
         You are a layout agent, and your task is to generate the bounding boxes for the objects. The following rules must be strictly followed during generation. A layout denotes a set of ``object: bounding box'' items. ``object'' means any object name, which starts the object name with “a” or “an” if possible.   ``bounding box'' is formulated as [x, y, w, h, d], where ``x, y'' denotes the top left coordinate of the bounding box, ``w'' denotes the width, ``h'' denotes the height, and ``d'' indicates the order of the object's front and back position in the image, starting from 0, the smaller d indicates the object is more forward. The top-left corner has coordinates [0, 0]. The bottom-right corner has coordinates [1, 1].  
          
        The output format: ``\{object$_1$: bounding box$_1$, object$_2$: bounding box$_2$, $\dots$, object$_n$: bounding box$_n$\}''

        \texttt{</SYS>} \\
        \midrule
        \texttt{<USER>} \\
        The input prompt is described as: \texttt{\{input prompt\}}.
        
        You should refer to the output of other agents: \texttt{\{outputs\}}.

        Please generate and layout according to the task description and rules.
        \\
        \texttt{</USER>[INST]}
        \\
        \bottomrule
    \end{tabular}
\end{table}

\noindent \textbf{Aesthetics enhancement agent.}
The task of the aesthetic enhancement agent is to perform the role of an aesthetic guide to optimize the description of the object characteristics, thus enhancing the artistic and aesthetic qualities of the image.
Its prompt template is displayed in \Cref{tb: Aesthetics enhancement agent}.

\begin{table}[h]
    \centering
    \caption{The prompt template for the layout agent.}
    \label{tb: Aesthetics enhancement agent}
    \begin{tabular}{@{}p{0.95\linewidth}@{}} 
        \toprule
        \textit{Aesthetics enhancement agent} \\ 
        \midrule
        \texttt{[INST]<SYS>} \\
        As an aesthetic enhancement agent, you serve as an aesthetic guide, refining the descriptions of an object’s characteristics to amplify its artistic and aesthetic appeal. This involves thoughtful consideration of composition, color balance, texture, and other key elements contributing to the image’s overall impact.
         
        \texttt{</SYS>} \\
        \midrule
        \texttt{<USER>} \\
        The input prompt is: \texttt{\{input prompt\}}.

        You should refer to the output of other agents: \texttt{\{outputs\}}.

        Please generate a glorified characterization as required.
        \\
        \texttt{</USER>[INST]}
        \\
        \bottomrule
    \end{tabular}
\end{table}

\section{Case Study of MCP}
\begin{figure*}[t]
\centering
\includegraphics[width=0.9\textwidth]{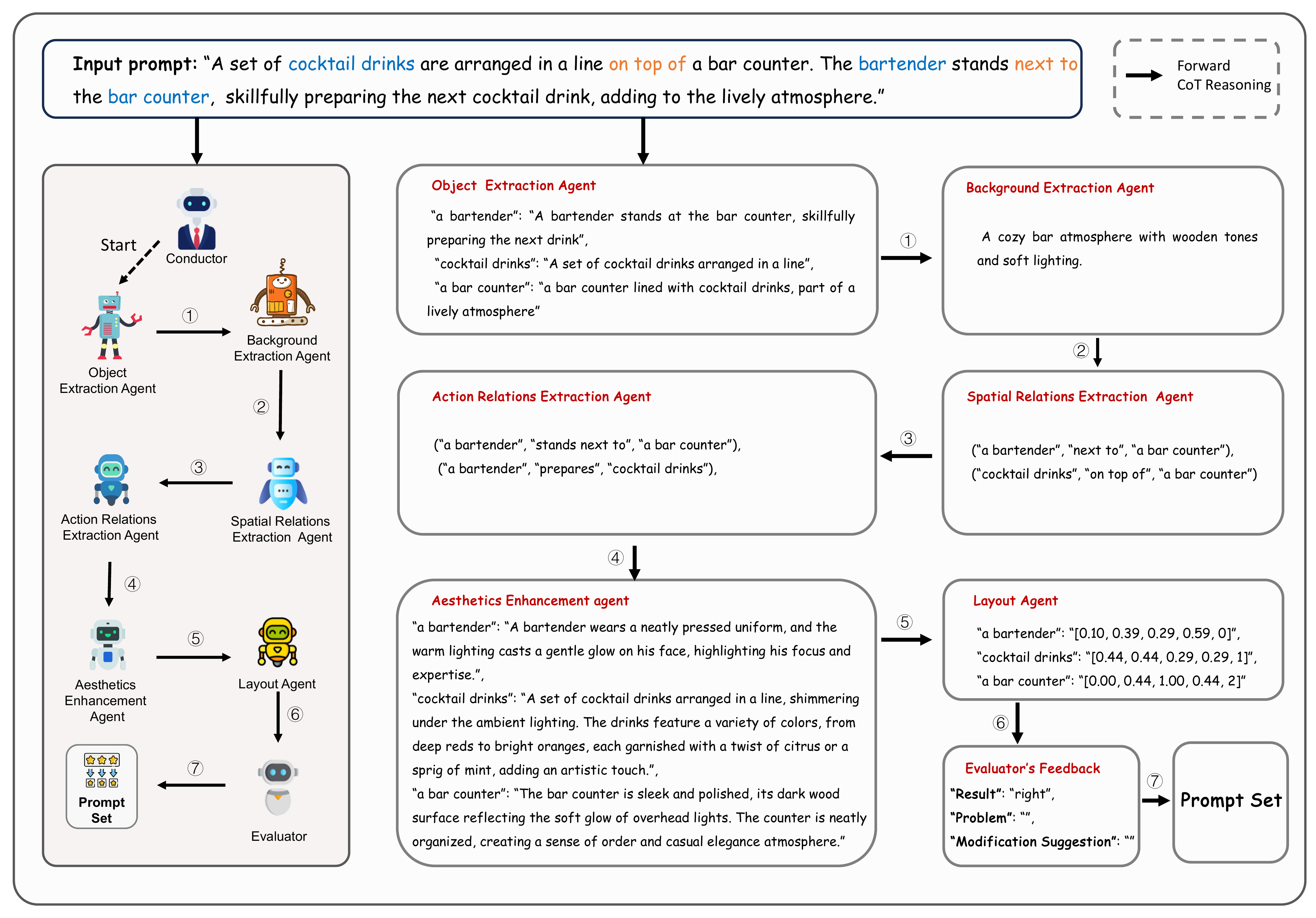}
\caption{A case to illustrate the workflow of MCP.}
\label{fig:supp_case_1}
\end{figure*}

\begin{figure*}[t]
\centering
\includegraphics[width=0.9\textwidth]{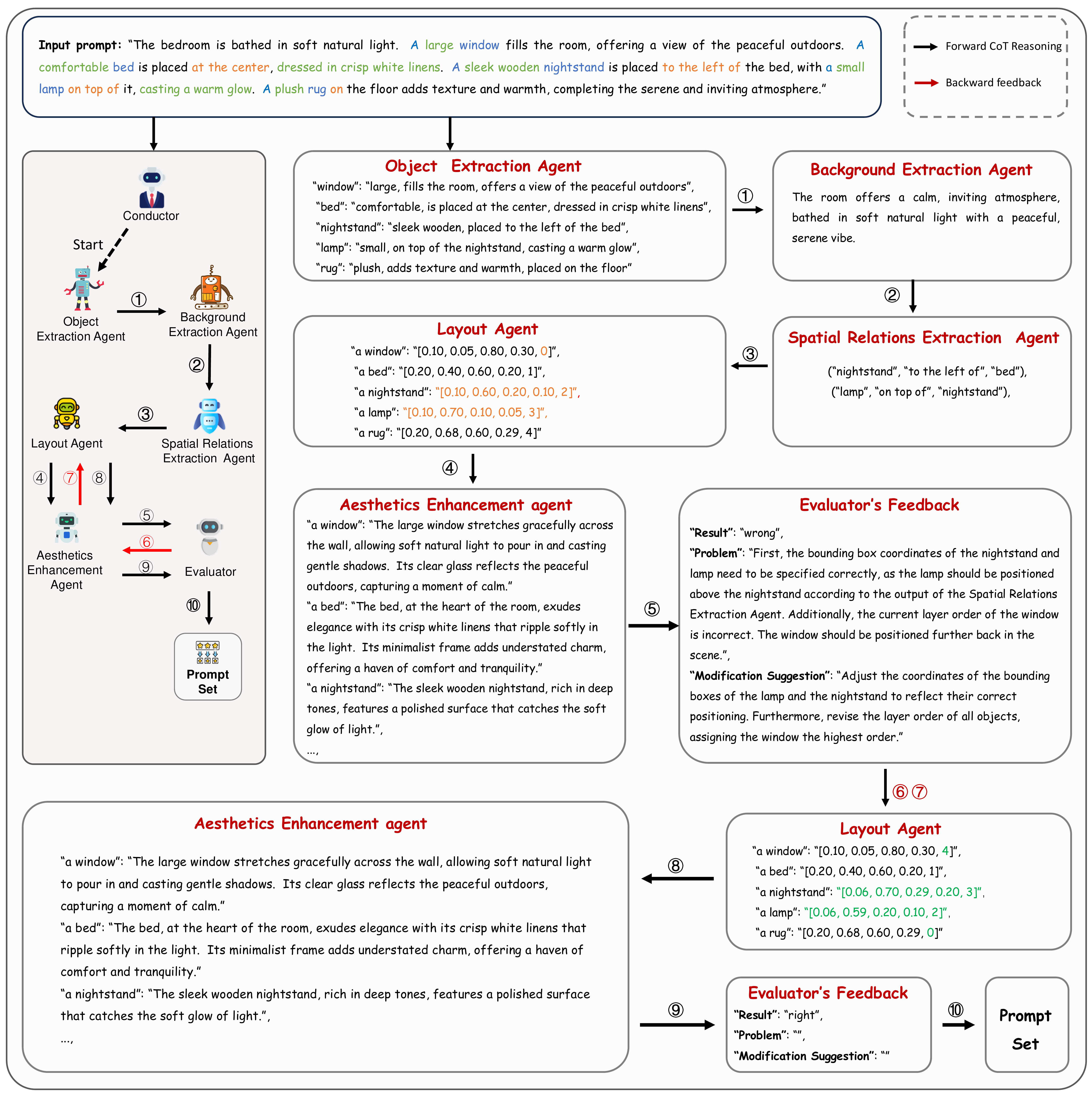}
\caption{A case to illustrate the workflow of MCP.  In the agents' outputs, orange text indicates errors, and green text indicates corrections. }
\label{fig:supp_case_2}
\end{figure*}

Figures \ref{fig:supp_case_1} and \ref{fig:supp_case_2}  show two cases of MCP workflows.
As shown in the figures, the conductor organizes multiple agents in an orderly manner according to the input prompt to achieve a well-collaborated result. \Cref{fig:supp_case_1} is a case where each agent correctly outputs the result. 
When a problem occurs during the system's execution of a task, the Evaluator can identify it through a backward feedback process and suggest modifications to resolve it quickly. 
For example, in the case of \Cref{fig:supp_case_2}, when the behavior of a certain agent does not meet the expectation, the Evaluator analyzes that the problem occurs in the layout agent and suggests a modification. The backward feedback process identifies the agent's error and regenerates the prompt set.
This mechanism of feedback and correction provides a high degree of flexibility and self-adaptation for the system, thus enhancing the robustness and intelligence of the whole system.

\section{Additional Ablation Results of MCP}
\begin{figure*}[t]
\centering
\includegraphics[width=0.85\textwidth]{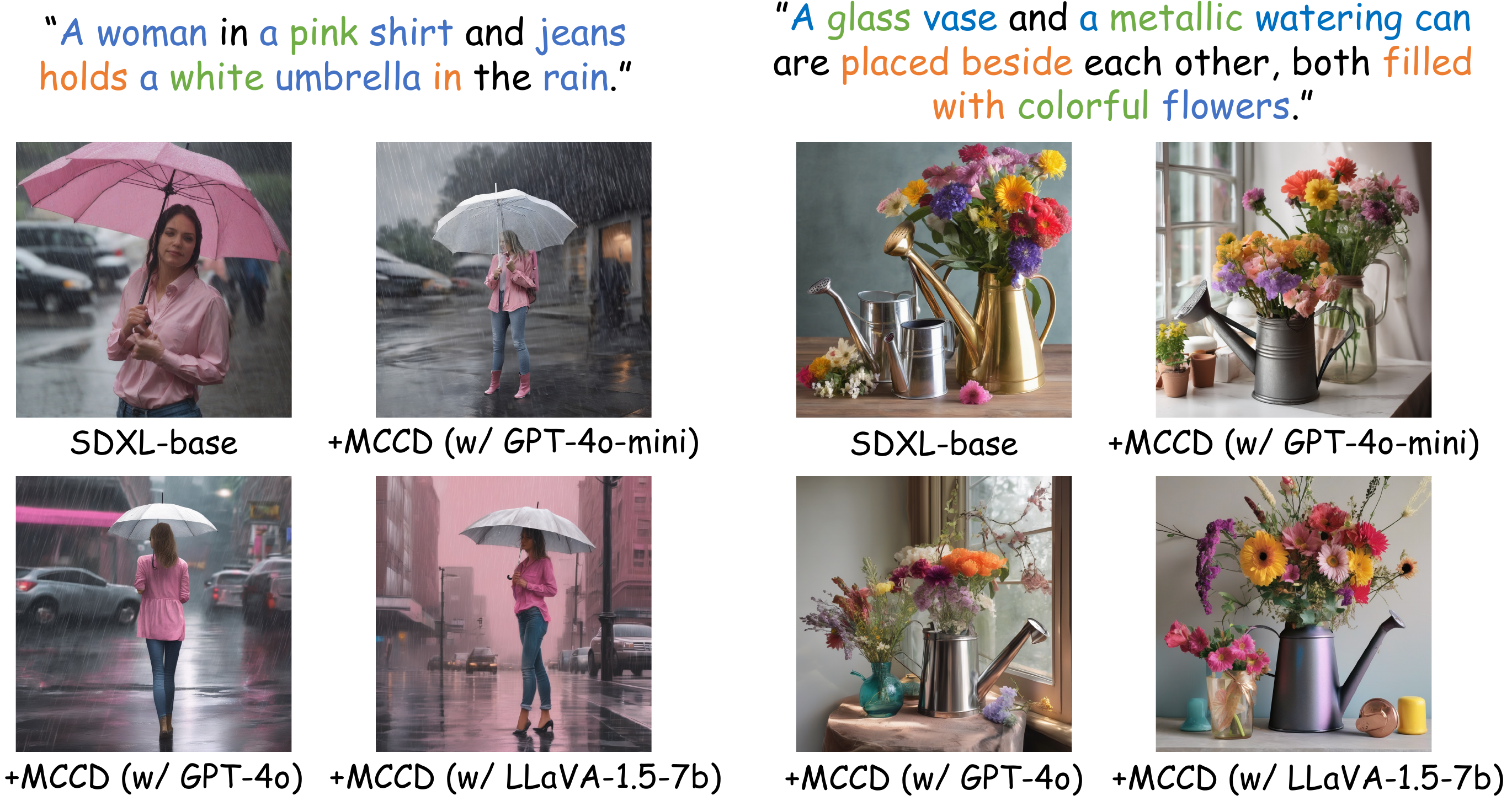}
\caption{Ablation results of MLLMs in MCP.}
\label{fig:supp_ablation_llm}
\end{figure*}

To demonstrate the scalability and generalizability of MCCD, we perform comprehensive ablation experiments on MLLMs used in MCP based on the SDXL-Base model. 
We select GPT-4o-mini \cite{achiam2023gpt}, GPT-4o \cite{achiam2023gpt}, and LLaVa-1.5-7b \cite{liu2024improved} as the MLLMs. The qualitative results are shown in \Cref{fig:supp_ablation_llm}. According to the experimental results, compared to the errors in object attribute binding and quantity generation of the SDXL-Base model, MCCD can generate correct image content with high fidelity and aesthetic quality after utilizing different MLLMs. Specifically, these models can accurately recognize and bind the attributes of objects while exhibiting high stability and consistency in the number and layout of objects, which significantly improves the quality and usability of the generated results. This suggests that MCCD can fully utilize the potential of these advanced MLLMs to generate images with high visual realism and artistic value in complex scenes.

\section{Additional Qualitative Analysis}
\begin{figure*}[t]
\centering
\includegraphics[width=1.0\textwidth]{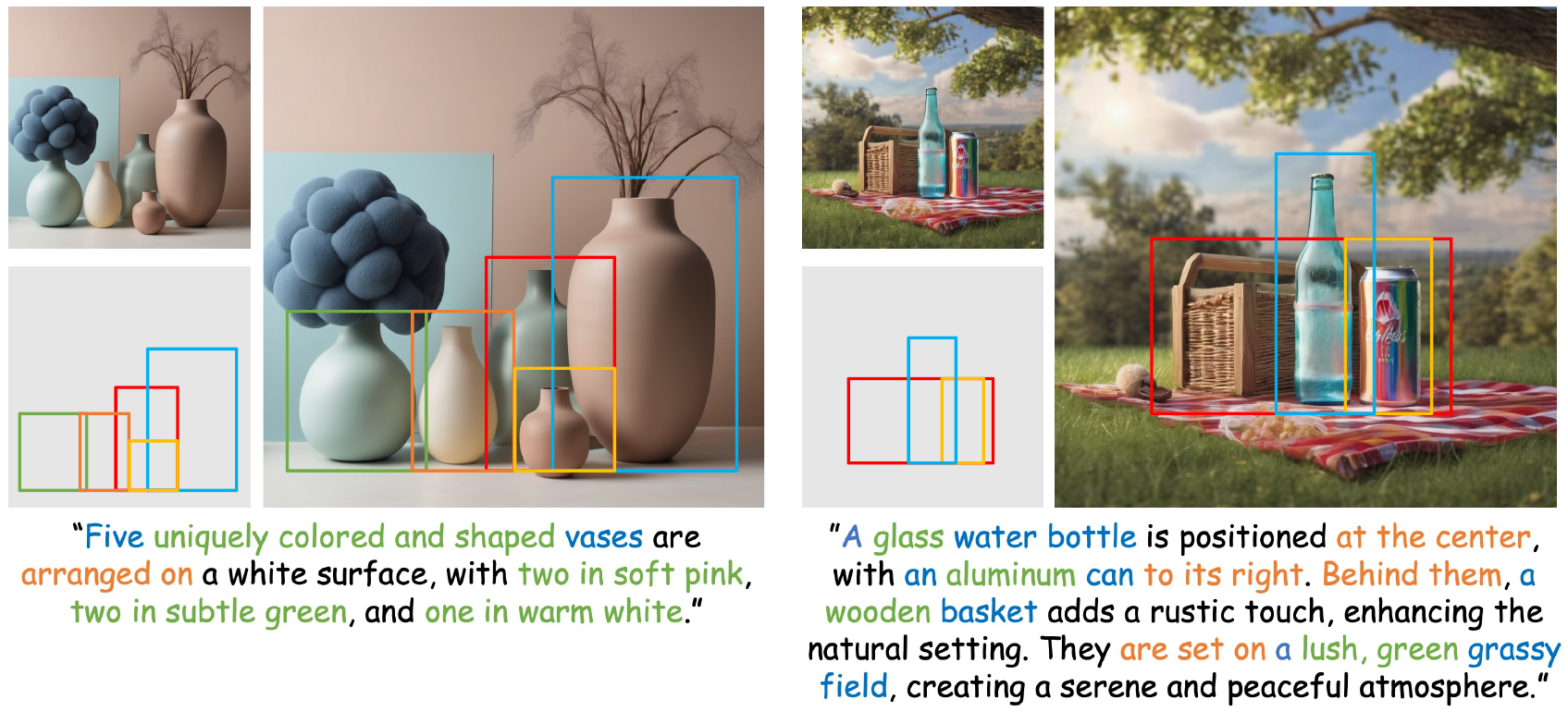}
\caption{Additional qualitative analysis.}
\label{fig: supp_vis}
\end{figure*}

\Cref{fig: supp_vis} shows the results of complex text prompts with multiple objects and attribute bindings by MCCD. Through an in-depth understanding and parsing of the input detailed text description, MCCD can effectively recognize and extract the objects and their interrelationships in the text, generating a reasonable and accurate bounding box layout. The position and size of each bounding box closely correspond to the elements (e.g., objects, colors, shapes, etc.) in the textual descriptions, ensuring the accurate positioning and realistic rendering of each element in the image. In addition, MCCD not only focuses on the physical positional relationship of the object but also fully captures the interaction between the object and its surroundings, resulting in an image that exhibits a high degree of realism and aesthetics. In this way, MCCD can transform complex and layered textual information into clearly structured and visually rich images, perfectly combining text and visual expression. MCCD can generate reasonable bounding box layouts, resulting in attractive images with high realism.

\section{Discussion of Broader Impact}
MCCD, as a training-free T2I approach with excellent performance, is capable of transforming complex and layered text information into clearly structured and visually rich images, perfectly combining text and visual representation. However, it also has potential negative impacts. For example, it may be used to generate scenarios involving immoral or illegal activities that can harm society. Additionally, automatically generated images may raise concerns about intellectual property and copyright.

\section{Discussion of Limitations and Future Work}
The proposed MCCD serves as a training-free plug-in that can be adapted to any Diffusion-based T2I methods, using a hierarchical compositional generative paradigm to enhance the quality of complex scene generation for the model. 
A potential problem is that the MCCD inference overhead is somewhat affected by the number of objects in a complex text prompt. As the number of bounding boxes increases, the inference time also increases. In the future, we will focus on designing efficient performance optimization strategies to improve the inference speed of the method.

\section*{Acknowledgements}
This work is supported in part by the Shanghai Municipal Science and Technology Committee of Shanghai Outstanding Academic Leaders Plan (No.\,21XD1430300), and in part by the National Key R\&D Program of China (No.\,2021ZD0113503).


{
    \small
    \bibliographystyle{ieeenat_fullname}
    \bibliography{main}
}

\end{document}